\documentclass[runningheads]{llncs}
\usepackage[T1]{fontenc}
\usepackage{graphicx}
\usepackage{verbatim}
\usepackage{multirow}
\usepackage{hyperref}
\usepackage[normalem]{ulem}
\useunder{\uline}{\ul}{}

\usepackage{array}
\begin{document}
\title{SurgX: Neuron-Concept Association for Explainable Surgical Phase Recognition}

\titlerunning{SurgX}
\author{Ka Young Kim \and
Hyeon Bae Kim \and
Seong Tae Kim\textsuperscript{${\ast}$}}

\authorrunning{Kim et al.}
%
\institute{Kyung Hee University, Yongin, Republic of Korea\\
\email{\{uwrgoy7584, hyeonbae.kim, st.kim\}@khu.ac.kr}\\}

\def\thefootnote{$\ast$}\footnotetext{Corresponding author}
\maketitle
\begin{abstract}
Surgical phase recognition plays a crucial role in surgical workflow analysis, enabling various applications such as surgical monitoring, skill assessment, and workflow optimization. Despite significant advancements in deep learning–based surgical phase recognition, these models remain inherently opaque, making it difficult to understand how they make decisions. This lack of interpretability hinders trust and makes it challenging to debug the model. To address this challenge, we propose SurgX, a novel concept-based explanation framework that enhances the interpretability of surgical phase recognition models by associating neurons with relevant concepts. In this paper, we introduce the process of selecting representative example sequences for neurons, constructing a concept set tailored to the surgical video dataset, associating neurons with concepts and identifying neurons crucial for predictions. Through extensive experiments on two surgical phase recognition models, we validate our method and analyze the explanation for prediction. This highlights the potential of our method in explaining surgical phase recognition. The code is available at \url{https://github.com/ailab-kyunghee/SurgX}.


\end{abstract}

\section{Introduction}
\label{introduction}
Surgical workflow analysis is crucial for computer-assisted surgery, as it enables surgical artificial intelligence systems to understand the sequential and temporal dynamics inherent in surgical procedures~\cite{kirtac2022surgical,maier2018surgical,padoy2012statistical}. Surgical phase recognition stands out as one of the critical tasks in surgical workflow analysis, classifying surgical video frames into their respective phase. 
Reflecting its clinical importance, a growing body of work has focused on developing advanced deep learning–based surgical phase recognition models~\cite{czempiel2020tecno,czempiel2021opera,liu2025lovit,liu2023skit,twinanda2016endonet}.

Although deep learning models have made remarkable progress in surgical phase recognition, they still encounter persistent challenges stemming from their inherent “black-box” nature, which refers to the difficulty in interpreting how these models process information and make decisions. 
The major challenge is the difficulty in trusting the model’s decisions. In surgical domains, models are required to have a transparent decision-making process in compliance with regulations like the European Union’s General Data Protection Regulation (GDPR)~\cite{salahuddin2022transparency,temme2017algorithms}, which emphasizes the need for sufficient explanations of model decisions.
If we cannot determine which concepts the model has learned and why it makes certain decisions, effectively improving the model and trusting its decisions becomes a significant challenge.
This challenge is further exacerbated in video-based models due to their inherent temporal dependency and complexity of models.

Previous studies on concept-based explainability for image-based models have shown that neurons in deep networks tend to align with human-interpretable concepts~\cite{ahn2024unified,bau2017network,kalibhat2023identifying,kim2024mask,khakzar2021towards}. One of the earliest approaches, Network Dissect~\cite{bau2017network} associates neurons with human-understandable concepts by computing the overlap between pixel-wise segmentation masks and the feature maps of individual neurons.  To address the scarcity of expensive segmentation mask datasets, Oikarinen \textit{et al.} proposed leveraging CLIP~\cite{radford2021learning}, a Vision-Language Model (VLM), to perform neuron-concept association~\cite{oikarinen2022clip}. In the medical domain, where segmentation mask datasets are even more limited,~\cite{kim2024mask} demonstrated that associating neurons with concepts is effective in interpreting medical diagnosis models. 

While explainability in image-based models has been extensively studied, its application to surgical video models remains largely underexplored, as their inherent temporal complexities introduce unique challenges in understanding decision-making processes. To bridge this gap, we propose \textbf{SurgX}, a novel concept-based explanation framework specifically designed for interpreting surgical phase recognition models with concepts. 
We introduce a systematic approach for selecting representative example sequences for each neuron, constructing concept sets, and identifying the most influential neurons in each prediction. We employ SurgX to interpret two surgical phase recognition models (TeCNO~\cite{czempiel2020tecno} and Causal ASFormer~\cite{yi2021asformer}) and evaluate our methods by examining whether the identified concept-neuron associations are meaningful and whether they contribute to explaining the model’s predictions.
Our main contributions are as follows:
\begin{itemize}
    \item We propose SurgX, a novel concept-based explanation framework, to interpret surgical phase recognition models. 
    To the best of our knowledge, this is the first study to provide a concept-based explanation for surgical phase recognition models.
    \item To ensure that concepts are appropriately associated with neurons in surgical phase recognition models, we introduce methods to construct the concept set specialized for a cholecystectomy video dataset. We then analyze how different concept sets can be used to explain model behavior more effectively, highlighting best practices for concept set selection in surgical applications.
    \item We validate SurgX on two surgical phase recognition models by examining whether the identified concept-neuron associations are meaningful and whether they contribute to explaining the model’s predictions.
\end{itemize}

\section{Methodology}
\label{methodology}

\subsection{Concept Set Construction}
\label{method_concept}
To ensure accurate neuron-concept annotation, it is essential to use a properly collected set of concepts that the neurons may have learned from the dataset.
We introduce three concept sets focused on cholecystectomy-related concepts.
First, we leverage the action triplets labels provided by the CholecT45~\cite{Nwoye_2022} dataset to build two concept sets, CholecT45-W and CholecT45-S. CholecT45-W comprises 30 words (e.g., grasper, clipping, gallbladder) derived from the action triplet labels in CholecT45. CholecT45-S contains 100 sentences created from the same action triplets following the prompting approach as in~\cite{yuan2023learning} (e.g., I use a hook to dissect the cystic plate). Lastly, the ChoLec-270 set was constructed by collecting a total of 270 concepts extracted from 11 cholecystectomy lecture videos available on the open surgical e-learning platform WebSurg, as well as from 4 cholecystectomy-related articles~\cite{Hassler2025,Kim2018,ALES5766,Strasberg2019}.

\subsection{Neuron Representative Sequence Selection}
\label{method_selection}
\begin{figure*}[t]
    \centering
    \includegraphics[width=\textwidth]{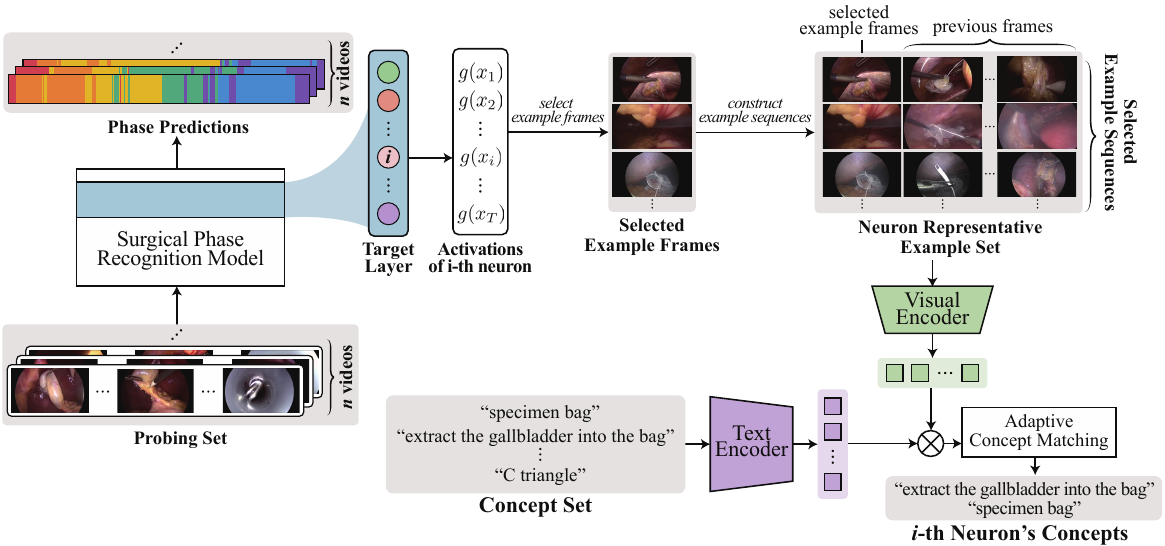}
    \caption{\small\textbf{Neuron concept annotation method for surgical phase recognition model.} 
    To interpret the neuron representations with concepts, neuron representative examples are selected, and a text-based concept set is prepared. The neuron representative examples are constructed based on the frames that are highly activated for the neuron in the video probing set. Both the example set and the concept set are projected into the vision-language model embedding space, where the neuron's concept is identified through a similarity-based concept matching method.
    }
    \label{fig:overall_concept_annotation} 
\end{figure*}

In this work, each neuron's concept association is determined by measuring the similarity between its representative sequences and candidate concepts, which underscores the importance of selecting sequences that accurately reflect each neuron's learned patterns. To this end, we follow a two-step process, illustrated in Figure~\ref{fig:overall_concept_annotation}. First, we select the example frames from the video probing set that highly activate each neuron. Since we use activations after the ReLU function, negative values are not considered. Second, we construct sequences by including \textit{n} previous frames with a dilation rate of \textit{m} from each selected example frame, considering that in surgical phase recognition models, the activation of a neuron at a given feature is influenced not only by the selected frame but also by previous frames. 

\subsection{Neuron-Concept Annotation}
\label{method_overall_annotation}
Figure~\ref{fig:overall_concept_annotation} shows an overview of the neuron concept annotation. 
Using the concept set constructed in Section~\ref{method_concept} and the neuron representative example set selected in Section~\ref{method_selection}, we perform concept annotation by following~\cite{ahn2024unified}. First, we extract text features from the concept set and visual features from neuron representative example set. Then, we compute the cosine similarity between the text and visual features.
\begin{equation}\label{equ:Calculate_cossim}
    s_{n,c} = \frac{1}{E} \sum_{e=1}^{E} \cos\bigl(v_{n,e}, t_{c}\bigr),
\end{equation}

\noindent where \( s_{n,c} \) denotes the concept score between \( n\)-th neuron and \( c \)-th concept which quantifies the extent to which a neuron represents a concept. 
For each of the \( E \) frames in the \( n \)-th neuron’s representative example set, we compute the cosine similarity between the visual feature \( v_{n,e} \) of \( e\)-th frame and the text feature \( t_{c} \) of \( c \)-th concept. 
The final score \( s_{n,c} \) is obtained by averaging these similarities across all selected frames. If \( s_{n,c} \) exceeds a threshold $\theta_{concept}$, \( c \)-th concept is annotated to \( n \)-th neuron. 

\begin{figure*}[t]
    \centering
    \includegraphics[width=0.9\textwidth]{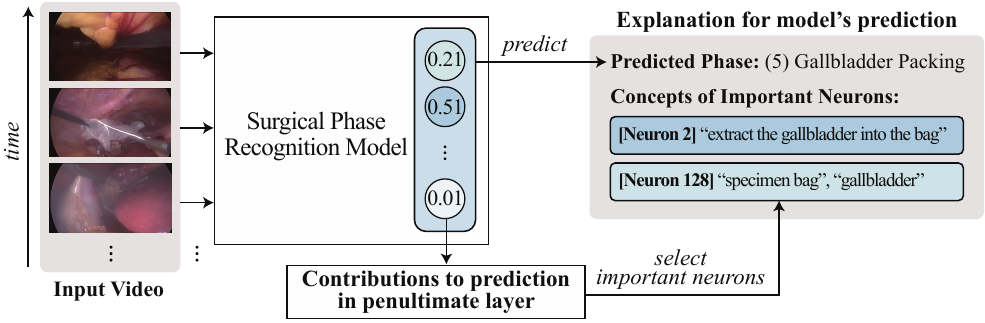}
    \caption{\small\textbf{Explanation of model's decisions.} 
    Important neurons are selected based on the contribution of the neurons in the penultimate layer to the model's prediction. The concepts annotated to these important neurons are used to explain the model decision.
     }
    \label{fig:overall_decision} 
\end{figure*}

\subsection{Explaining Model's Decision using Highly Contributed Neurons}

After annotating each neuron with concepts, the model's predictions can be explained based on the concepts of important neurons that significantly contributed to its predictions.
The process of explaining a prediction is illustrated in Figure \ref{fig:overall_decision}. 
When a surgical phase recognition model predicts the phase of a specific frame, various neurons in the penultimate layer, each representing different concepts, contribute to the prediction with different extents. Inspired by \cite{khakzar2021neural,ahn2023lineoutofdistributiondetectionleveraging}, we first calculate the contribution of each neuron to the final prediction to explain the model decisions as follows:
%
\begin{equation}\label{equ:contribution}
    c_{p,i}(x) = \left| f_\theta(x) - f_\theta(x;a_{p,i}\leftarrow 0)\right| = \left| a_{p,i}\nabla_{a_{p,i}}f_\theta(x) \right|,
\end{equation}
\noindent where \( f_\theta(x) \) is the prediction of model with parameter $\theta$ for input \( x \), and \( a_{p,i} \) is the activation of the \( i \)-th neuron for class \( p \). The term \( f_\theta(x; a_{p,i} \leftarrow 0) \) represents the prediction without this neuron’s influence, making \( | f_\theta(x) - f_\theta(x; a_{p,i} \leftarrow 0) | \) be a measure of its contribution.
Then, we identify the neurons that had a significant impact on the prediction. 
By utilizing the concepts matched to these highly contributing neurons, interpretable explanations of why the model made a particular decision can be provided.

\section{Experiments}
\label{experiments}

\subsection{Experimental Setup}
\noindent \textbf{Model.}
As most surgical phase recognition models rely on either TCN or Transformer architectures, we select TeCNO~\cite{czempiel2020tecno} and Causal ASFormer~\cite{yi2021asformer} as target models to interpret and highlight the generality of SurgX. 
TeCNO is a TCN-based model commonly used for surgical phase recognition.
On the other hand, ASFormer is a Transformer-based model widely used for temporal action segmentation in the computer vision domain.
To reflect real-world clinical settings, we modify ASFormer to operate causally using only current and past frames, and we refer to it as Causal ASFormer in this paper. LoViT~\cite{liu2025lovit} features are used as the spatial features for Causal ASFormer. Additionally, for extracting visual features from neuron representative sequences and text features from concepts, we employ SurgVLP~\cite{yuan2023learning}, a vision-language model specifically trained for surgical domain. Using the visual encoder of SurgVLP to extract visual features from representative frames avoids the gap between the embedding spaces of the phase recognition model and SurgVLP.

\noindent \textbf{Dataset.}
We employ the Cholec80 dataset~\cite{twinanda2016endonet}, which comprises 80 laparoscopic cholecystectomy videos annotated with surgical phases. Following previous works~\cite{gao2021trans,jin2017sv}, we split the dataset into 40 training videos and 40 test videos. The training set is used for both model training and probing set in Section~\ref{method_selection}, while the test set is reserved for evaluating the interpretability of SurgX.

\noindent \textbf{Evaluation Metrics.}
To quantitatively evaluate SurgX, we are inspired by the evaluation metrics in~\cite{ahn2024unified,oikarinen2022clip,kim2024mask} and evaluate the interpretability of our approach based on two key aspects.
First, we define Concept Alignment Score to evaluate how well concepts are annotated to neurons by measuring the cosine similarity between the concepts assigned to neurons in the final layer and the ground-truth class as in~\cite{ahn2024unified,oikarinen2022clip,kim2024mask}.
Second, we define Prediction Interpretability Score to evaluate how well the model's decisions can be explained by measuring the cosine similarity between the model's prediction and the concepts annotated to highly contributing neurons in the penultimate layer.
For both aspects, we compute the cosine similarity between the concept and the phase expressed in word form, as well as between the concept and the phase expressed in sentence form provided in~\cite{yuan2023learning}, and then average the two scores, since all the concepts and frames are encoded by SurgVLP~\cite{yuan2023learning} which can be influenced by the text format.

\subsection{Quantitative Evaluation of SurgX}
This section presents the ablation studies for each part introduced in Section~\ref{methodology}, along with the quantitative results of SurgX.
The ablation studies were conducted on Causal ASFomer. The Concept Alignment Score was measured on the output of the last Conv1d layer, the final layer, where the output dimension corresponds to the number of phases (i.e., 7). 
Additionally, the prediction interpretability score was measured on the output of the last transformer block, the penultimate layer, where the output dimension is 256. 
\begin{table}[t]
\caption{Analysis of concept set construction methods.}
\centering
{
\begin{tabular}{c|>{\centering}p{1.1cm}>{\centering}p{1.1cm}c|>{\centering}p{1.1cm}>{\centering}p{1.1cm}c|c}
\hline
\multirow{2}{*}{Concept Set} & \multicolumn{3}{c|}{Concept Alignment } & \multicolumn{3}{c|}{Prediction Interpretability}  &Unique    \\ 
             & Word     & Sentence   & Avg      & Word   & Sentence & Avg     & Concept \\ \hline
CholecT45-W  & 0.4078   & 0.3681     & \underline{0.3880}   & 0.5978 & 0.5099   & \underline{0.5539}  & 7       \\
CholecT45-S  & 0.3000   & 0.4538     & 0.3769   & 0.4679 & 0.5564   & 0.5122  & \underline{11}      \\
ChoLec-270    & 0.4153   & 0.4796     & \textbf{0.4475}   & 0.6532 & 0.5452   & \textbf{0.5992}  & \textbf{34}      \\ \hline
\end{tabular}
}
\label{tab:conceptset}
\end{table}

\noindent \textbf{Concept Set Construction.}
To evaluate the concept sets, we conducted experiments while keeping other settings fixed as Video-wise Threshold Selection and Dilated Sequences Selection with a 5-second interval. 
Table \ref{tab:conceptset} shows that the Concept Alignment Score and the Prediction Interpretability Score of ChoLec-270 were the highest, demonstrating that using ChoLec-270 explains the model best among the compared concept sets.

\begin{table}[t]
\centering
\caption{Ablation study on the neuron representative frame selection methods.}
{
\begin{tabular}{c|>{\centering}p{1.1cm}>{\centering}p{1.1cm}c|>{\centering}p{1.1cm}>{\centering}p{1.1cm}c}
\hline
\multirow{2}{*}{Method}  
                        & \multicolumn{3}{c|}{Concept Alignment}                & \multicolumn{3}{c}{Prediction Interpretability} \\
                        & Word    & Sentence  & Avg        & Word   & Sentence & Avg     \\ \hline
Global Threshold        & 0.3222  &  0.4621   & 0.3922 & 0.5586 & 0.4819   & 0.5203  \\
Global TopK             & 0.3099  &  0.4622   & 0.3861 & 0.5369 & 0.4531   & 0.4950  \\
Video-wise Threshold         & 0.4153  &  0.4796   & \underline{0.4475}  & 0.6532 & 0.5452   & \textbf{0.5992}  \\
Video-wise TopK             & 0.4146  &  0.4824   & \textbf{0.4485}  & 0.6454 & 0.5403   & \underline{0.5929}  \\ \hline
\end{tabular}
\label{tab:frame_selection}
}
\end{table}
\noindent \textbf{Neuron Representative Frame Selection.}
This study is conducted under two settings: \textbf{Global setting}, where representative frames are selected from all video frames, and \textbf{Video-wise setting}, where frames are first selected for each video and then aggregated. 
In each setting, two selection approaches are applied: (1) selecting the top-K frames based on activation values (K=40 for the global setting and K=1 for the video-wise setting) and (2) applying an adaptive thresholding method ($\alpha$=0.95) from Adaptive Neuron Representative Image Selection in~\cite{kim2024mask}. Table \ref{tab:frame_selection} shows that the Video-wise Threshold method, which first selects frames for each video using an adaptive threshold before aggregation, achieves the best performance. 
In the global setting, frames are often concentrated in a few specific videos and include many adjacent frames, whereas the video-wise setting selects more diverse frames across videos, enabling the model to capture common concepts across different videos more effectively.

\begin{table}[t]
\centering
\caption{Ablation study on the neuron representative sequence selection methods using cosine similarity. The number in parentheses (\#) denotes the dilation size.}
{
\begin{tabular}{c|>{\centering}p{1.1cm}>{\centering}p{1.1cm}c|>{\centering}p{1.1cm}>{\centering}p{1.1cm}c}
\hline
\multirow{2}{*}{Method}  
                        & \multicolumn{3}{c|}{Concept Alignment}   & \multicolumn{3}{c}{Prediction Interpretability} \\
                      & Word   & Sentence & Avg        & Word   & Sentence & Avg \\ \hline
Single-Frame          & 0.3576 & 0.4569   & 0.4073  & 0.6426 & 0.5513   & 0.5970    \\
Contiguous-Sequence   & 0.3641 & 0.4711   & 0.4176  & 0.6458 & 0.5487   & 0.5973    \\
Dilated-Sequence (3)  & 0.3534 & 0.4920   & \underline{0.4227} & 0.6545 & 0.5499   & \underline{0.6022}    \\
Dilated-Sequence (5)  & 0.4153 & 0.4796   & \textbf{0.4475} & 0.6532 & 0.5452   & 0.5992    \\
Dilated-Sequence (10) & 0.3684 & 0.4004   & 0.3844 & 0.6856 & 0.5548   & \textbf{0.6202}    \\ \hline
\end{tabular}
}
\label{tab:sequence_selection}
\end{table}
\noindent \textbf{Representative-Frame-Based Sequence Construction.}
To consider the temporal attributes of the Surgical Phase Recognition model, we conducted an ablation study to construct a neuron representative frame sequence. 
Table \ref{tab:sequence_selection} presents a comparison of performance between using a single frame and a sequence of 10 frames.
The sequence of 10 frames was configured to include frames selected at 1-second intervals (Contiguous-Sequence) as well as frames selected at 3-second, 5-second, and 10-second intervals (Dilated-Sequence (3), Dilated-Sequence (5), and Dilated-Sequence (10), respectively). The results demonstrated that using a sequence of frames outperform annotating concepts with a single frame.
Additionally, the application of the dilation technique, which enables the sequence to incorporate broader contextual information from preceding frames based on the representative frame, is found to improve performance.

\begin{figure*}[t]
    \centering
    \includegraphics[width=0.95\textwidth]{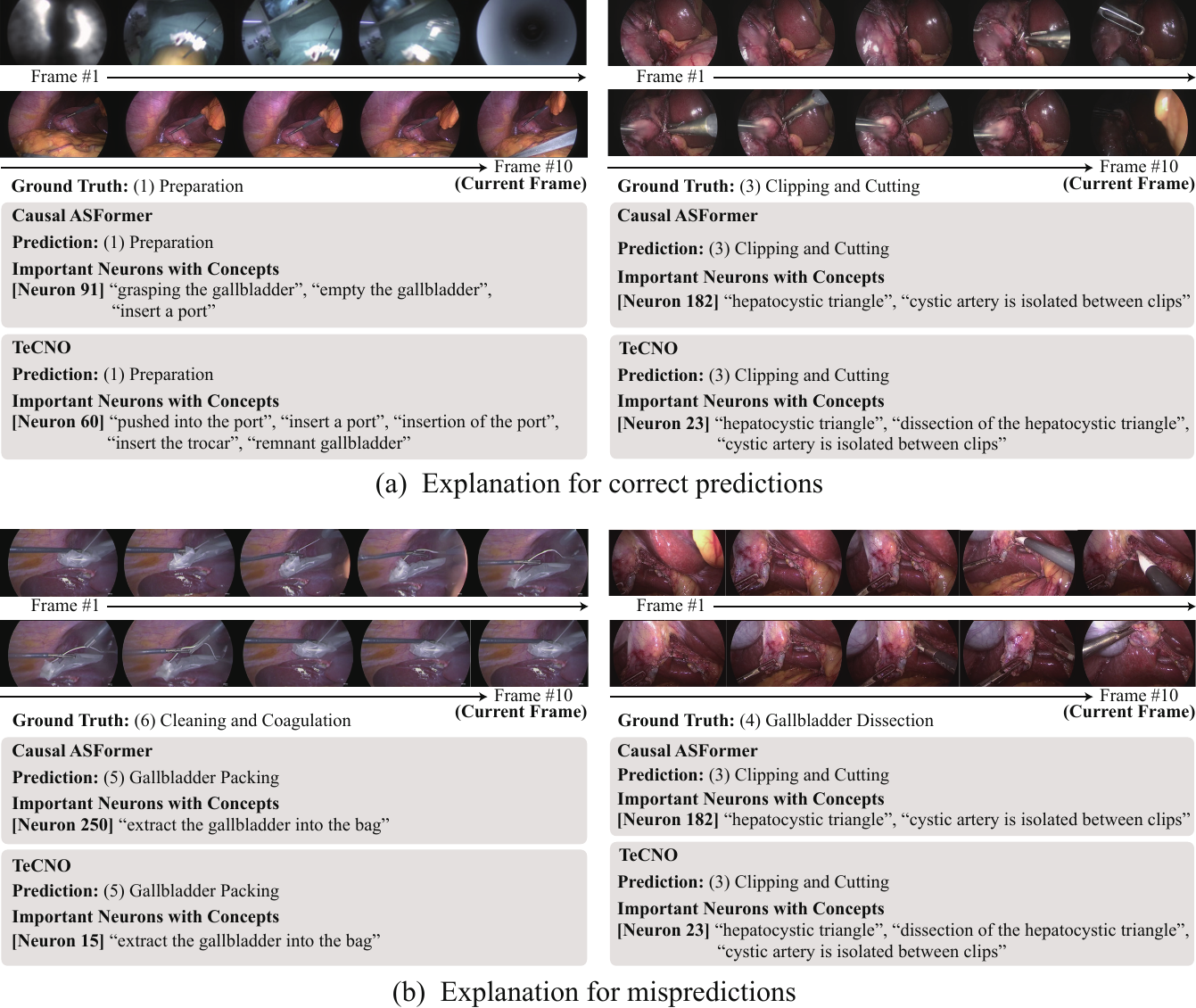}
    \caption{\small\textbf{Qualitative results of SurgX.} 
Predictions of Causal ASFormer and TeCNO for the current frame, along with the concepts of the neurons that strongly contribute to those predictions. Only the best-contributing neuron is shown for each explanation.
    }
    \label{fig:qual_modelwise} 
\end{figure*}

\subsection{Discussion}
Figure~\ref{fig:qual_modelwise} shows examples of explanations for predictions made by Causal ASFormer and TeCNO, visualizing the best-contributing neuron to the predictions and the corresponding concepts annotated to the neuron. We analyze the reason for the model's predictions based on the explanation examples.

\noindent \textbf{Analysis of Correct Prediction.}
Figure \ref{fig:qual_modelwise} (a) illustrates the concepts of the neurons that contribute most to correct predictions. In the left example of (a), although the current frame does not contain a port, both Causal ASFormer and TeCNO incorporate information from previous frames, allowing neurons associated with ``Insert a port'' and ``Pushed into the port'' to play a crucial role in accurately predicting the ``(1) Preparation'' phase.
In the right example of (a), despite the darkness of the current frame, neurons associated with ``hepatocystic triangle'' and ``cystic artery is isolated between clips'' in both models contribute the most to accurately predicting the ``(3) Clipping and Cutting''.
These examples demonstrate that Causal ASFormer and TeCNO do not rely solely on the current frame but instead leverage information from previous frames to learn and utilize relevant concepts for accurate predictions.

\noindent \textbf{Analysis of Incorrect Prediction.}
Figure~\ref{fig:qual_modelwise} (b) shows that SurgX also enables interpreting incorrect predictions.
In the left example of (b), a neuron annotated with ``extract the gallbladder into the bag'' misleads the model to predict ``(5) Gallbladder Packing'', while the ground truth phase is ``(6) Cleaning and Coagulation''. In the right example of (b), the presence of visible clips leads the model to detect the concept "cystic artery is isolated between clips," resulting in the misprediction ``(3) Clipping and Cutting''. When the GT phase is “(4) Gallbladder Dissection,” mispredictions as “(3) Clipping and Cutting” frequently occur in the models. In these cases, 88.22\% involves neurons associated with ``cystic artery is isolated between clips'' In contrast, in correctly predicted cases, 92.88\% do not involve these neurons.

\section{Conclusion}
\label{conclusion}
In this study, we propose SurgX, a concept-based explanation framework for surgical phase recognition. SurgX includes concept construction, neuron representative sequence selection, and concept annotation methods designed specifically for surgical phase recognition models.
Experiments on two surgical phase recognition models using the Cholec80 dataset demonstrate that SurgX provides meaningful concept-neuron associations, improving model explainability. By offering insights into model predictions with surgical terminology-based concepts, SurgX enhances the transparency and reliability of surgical AI models.

\subsubsection{Acknowledgements.} 
This work was partly supported by the National Research Foundation of Korea (NRF) grant funded by the Korea government (MSIT) (No. RS-2024-00334321), by the Institute of Information \& Communications Technology Planning and Evaluation(IITP) grant funded by the Korea government (MSIT) (No. RS-2022-II220078, No. RS-2024-00509257, No. RS-2022-00155911) and ITRC(Information Technology Research Center) grant (IITP-2025-RS-2023-00258649).

\subsubsection{\discintname}
The authors have no competing interests to declare that are relevant to the content of this article. 

\bibliographystyle{splncs04}
\bibliography{main}

\begin{thebibliography}{10}
\providecommand{\url}[1]{\texttt{#1}}
\providecommand{\urlprefix}{URL }
\providecommand{\doi}[1]{https://doi.org/#1}

\bibitem{ahn2024unified}
Ahn, Y.H., Kim, H.B., Kim, S.T.: Www: A unified framework for explaining what where and why of neural networks by interpretation of neuron concepts. In: CVPR. pp. 10968--10977 (2024)

\bibitem{ahn2023lineoutofdistributiondetectionleveraging}
Ahn, Y.H., Park, G.M., Kim, S.T.: Line: Out-of-distribution detection by leveraging important neurons (2023), \url{https://arxiv.org/abs/2303.13995}

\bibitem{bau2017network}
Bau, D., Zhou, B., Khosla, A., Oliva, A., Torralba, A.: Network dissection: Quantifying interpretability of deep visual representations. In: Proceedings of the IEEE conference on computer vision and pattern recognition. pp. 6541--6549 (2017)

\bibitem{czempiel2020tecno}
Czempiel, T., Paschali, M., Keicher, M., Simson, W., Feussner, H., Kim, S.T., Navab, N.: Tecno: Surgical phase recognition with multi-stage temporal convolutional networks. In: International Conference on Medical Image Computing and Computer Assisted Intervention. pp. 343--352 (2020)

\bibitem{czempiel2021opera}
Czempiel, T., Paschali, M., Ostler, D., Kim, S.T., Busam, B., Navab, N.: Opera: Attention-regularized transformers for surgical phase recognition. In: International Conference on Medical Image Computing and Computer Assisted Intervention. pp. 604--614 (2021)

\bibitem{gao2021trans}
Gao, X., Jin, Y., Long, Y., Dou, Q., Heng, P.A.: Trans-svnet: Accurate phase recognition from surgical videos via hybrid embedding aggregation transformer. In: International Conference on Medical Image Computing and Computer Assisted Intervention. pp. 593--603 (2021)

\bibitem{Hassler2025}
Hassler, K., Collins, J., Philip, K., Jones, M.: Laparoscopic Cholecystectomy. StatPearls Publishing (2025)

\bibitem{jin2017sv}
Jin, Y., Dou, Q., Chen, H., Yu, L., Qin, J., Fu, C.W., Heng, P.A.: Sv-rcnet: workflow recognition from surgical videos using recurrent convolutional network. IEEE transactions on medical imaging  \textbf{37}(5),  1114--1126 (2017)

\bibitem{kalibhat2023identifying}
Kalibhat, N., Bhardwaj, S., Bruss, C.B., Firooz, H., Sanjabi, M., Feizi, S.: Identifying interpretable subspaces in image representations. In: International Conference on Machine Learning. pp. 15623--15638 (2023)

\bibitem{khakzar2021neural}
Khakzar, A., Baselizadeh, S., Khanduja, S., Rupprecht, C., Kim, S.T., Navab, N.: Neural response interpretation through the lens of critical pathways. In: CVPR. pp. 13528--13538 (2021)

\bibitem{khakzar2021towards}
Khakzar, A., Musatian, S., Buchberger, J., Valeriano~Quiroz, I., Pinger, N., Baselizadeh, S., Kim, S.T., Navab, N.: Towards semantic interpretation of thoracic disease and covid-19 diagnosis models. In: International Conference on Medical Image Computing and Computer Assisted Intervention. pp. 499--508 (2021)

\bibitem{kim2024mask}
Kim, H.B., Ahn, Y.H., Kim, S.T.: Mask-free neuron concept annotation for interpreting neural networks in medical domain. In: International Conference on Medical Image Computing and Computer-Assisted Intervention. pp. 524--533 (2024)

\bibitem{Kim2018}
Kim, S., Donahue, T.: Laparoscopic cholecystectomy. JAMA  \textbf{319}(17), ~1834 (2018)

\bibitem{kirtac2022surgical}
Kirtac, K., Aydin, N., Lavanchy, J.L., Beldi, G., Smit, M., Woods, M.S., Aspart, F.: Surgical phase recognition: From public datasets to real-world data. Applied Sciences  \textbf{12}(17), ~8746 (2022)

\bibitem{liu2025lovit}
Liu, Y., Boels, M., Garcia-Peraza-Herrera, L.C., Vercauteren, T., Dasgupta, P., Granados, A., Ourselin, S.: Lovit: Long video transformer for surgical phase recognition. Medical Image Analysis  \textbf{99},  103366 (2025)

\bibitem{liu2023skit}
Liu, Y., Huo, J., Peng, J., Sparks, R., Dasgupta, P., Granados, A., Ourselin, S.: Skit: a fast key information video transformer for online surgical phase recognition. IEEE/CVF International Conference on Computer Vision pp. 21074--21084 (2023)

\bibitem{maier2018surgical}
Maier-Hein, L., Eisenmann, M., Feldmann, C., Feussner, H., Forestier, G., Giannarou, S., Gibaud, B., Hager, G.D., Hashizume, M., Katic, D., et~al.: Surgical data science: A consensus perspective. arXiv preprint arXiv:1806.03184  (2018)

\bibitem{ALES5766}
Majumder, A., Altieri, M.S., Brunt, L.M.: How do i do it: laparoscopic cholecystectomy. Annals of Laparoscopic and Endoscopic Surgery  \textbf{5} (2020)

\bibitem{Nwoye_2022}
Nwoye, C.I., Yu, T., Gonzalez, C., Seeliger, B., Mascagni, P., Mutter, D., Marescaux, J., Padoy, N.: Rendezvous: Attention mechanisms for the recognition of surgical action triplets in endoscopic videos. Medical Image Analysis  \textbf{78},  102433 (2022)

\bibitem{oikarinen2022clip}
Oikarinen, T., Weng, T.W.: Clip-dissect: Automatic description of neuron representations in deep vision networks. ICLR  (2023)

\bibitem{padoy2012statistical}
Padoy, N., Blum, T., Ahmadi, S.A., Feussner, H., Berger, M.O., Navab, N.: Statistical modeling and recognition of surgical workflow. Medical image analysis  \textbf{16}(3),  632--641 (2012)

\bibitem{radford2021learning}
Radford, A., Kim, J.W., Hallacy, C., Ramesh, A., Goh, G., Agarwal, S., Sastry, G., Askell, A., Mishkin, P., Clark, J., et~al.: Learning transferable visual models from natural language supervision. In: International conference on machine learning. pp. 8748--8763 (2021)

\bibitem{salahuddin2022transparency}
Salahuddin, Z., Woodruff, H.C., Chatterjee, A., Lambin, P.: Transparency of deep neural networks for medical image analysis: A review of interpretability methods. Computers in biology and medicine  \textbf{140},  105111 (2022)

\bibitem{Strasberg2019}
Strasberg, S.M.: A three-step conceptual roadmap for avoiding bile duct injury in laparoscopic cholecystectomy: an invited perspective review. J Hepatobiliary Pancreat Sci  \textbf{26}(4),  123--127 (2019)

\bibitem{temme2017algorithms}
Temme, M.: Algorithms and transparency in view of the new general data protection regulation. Eur. Data Prot. L. Rev.  \textbf{3}, ~473 (2017)

\bibitem{twinanda2016endonet}
Twinanda, A.P., Shehata, S., Mutter, D., Marescaux, J., De~Mathelin, M., Padoy, N.: Endonet: a deep architecture for recognition tasks on laparoscopic videos. IEEE transactions on medical imaging  \textbf{36}(1),  86--97 (2016)

\bibitem{yi2021asformer}
Yi, F., Wen, H., Jiang, T.: Asformer: Transformer for action segmentation. British Machine Vision Conference (BMVC)  (2021)

\bibitem{yuan2023learning}
Yuan, K., Srivastav, V., Yu, T., Lavanchy, J.L., Mascagni, P., Navab, N., Padoy, N.: Learning multi-modal representations by watching hundreds of surgical video lectures. arXiv preprint arXiv:2307.15220  (2023)

\end{thebibliography}

\end{document}